%% file: main.tex
\documentclass{article}
\usepackage{spconf,amsmath,graphicx}
\usepackage{makecell}

\usepackage{hyperref} 

\title{Augmenting Automatic Speech Recognition Models with Disfluency Detection}
\name{Robin Amann, Zhaolin Li, Barbara Bruno, Jan Niehues}
\address{Karlsruhe Institute of Technology, Germany}

\begin{document}
%
\maketitle
\begin{abstract}

Speech disfluency commonly occurs in conversational and spontaneous speech. However, standard Automatic Speech Recognition (ASR) models struggle to accurately recognize these disfluencies because they are typically trained on fluent transcripts. Current research mainly focuses on detecting disfluencies within transcripts, overlooking their exact location and duration in the speech. Additionally, previous work often requires model fine-tuning and addresses limited types of disfluencies. 

In this work, we present an inference-only approach to augment any ASR model with the ability to detect open-set disfluencies. We first demonstrate that ASR models have difficulty transcribing speech disfluencies. Next, this work proposes a modified Connectionist Temporal Classification(CTC)-based forced alignment algorithm from \cite{kurzinger2020ctc} to predict word-level timestamps while effectively capturing disfluent speech. Additionally, we develop a model to classify alignment gaps between timestamps as either containing disfluent speech or silence. This model achieves an accuracy of 81.62\% and an F1-score of 80.07\%. We test the augmentation pipeline of alignment gap detection and classification on a disfluent dataset. Our results show that we captured 74.13\% of the words that were initially missed by the transcription, demonstrating the potential of this pipeline for downstream tasks.
\end{abstract}
\begin{keywords}
Automatic speech recognition, speech disfluency, forced alignment
\end{keywords}

\input{sections/s1_intro}

\input{sections/s2_disfluency_detection}
\input{sections/s3_dataset}

\input{sections/s4_exp_results}

\input{sections/s5_conclusion}



\section{ACKNOWLEDGMENTS}
This work was funded by the Baden-Württemberg Ministry of Science, Research and Art (MWK), via the state digitalisation strategy digital@bw.


\bibliographystyle{IEEEbib}
\bibliography{main}

\end{document}

%% file: sections/s1_intro.tex
\section{Introduction}
\label{sec:intro}


Speech disfluency refers to interruptions in the flow of speech, such as repetitions, interjections, and revisions. It is a natural part of conversational and spontaneous speech but can be particularly pronounced and frequent in certain speech disorders, such as stuttering \cite{wu2023world, choo2023rate, kouzelis2023weaklysupervised}. Analyzing speech disfluency can aid in diagnosing speech disorders. It can also help in understanding language proficiency that can be applied, for example, in interviews and children's education. 

Manually annotating speech disfluency for analysis is costly, and Automatic Speech Recognition (ASR) can support the annotation process. The ASR systems transcribe the speech into readable text, which can then be passed to the evaluator or automatic evaluation pipelines for analysis. However, ASR models show performance degradation in disfluent speech, because the models are developed to generate fluent transcripts to enhance readability \cite{wu2023world, lea2023user}.


\begin{figure}[!h]
    \centering
    \includegraphics[width=8.5cm]{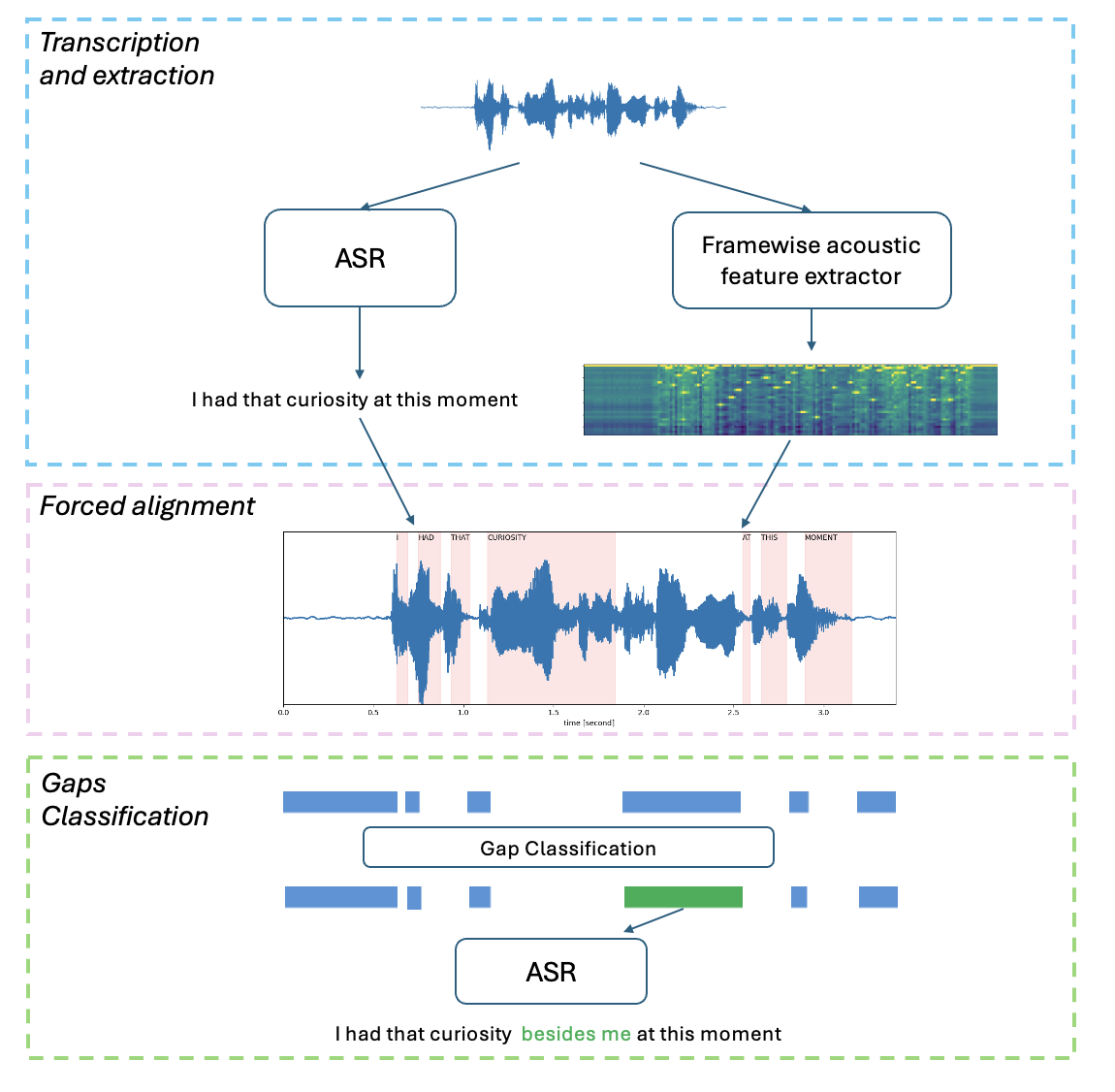}
    \caption{The pipeline to augment ASR models with disfluency detection with a follow-up re-transcription for example of application.}
    \label{fig:Approach}
\end{figure}

To detect speech disfluency with ASR models, one popular approach is post-processing the ASR predictions as a sequence labelling problem \cite{chen2022teaching, rocholl2021disfluency, rohanian-hough-2021-best, 9746067, mihajlik24_interspeech}. Alternatively, \cite{9528931, futami2023streaming, horii2022end, lian2023unconstrained, 10095555, mujtaba2024inclusive} focus on jointly predicting transcriptions and disfluencies with end-to-end speech recognition. In addition, \cite{Ma_2023} explores adapting the ASR foundation models, which are robust to recognize unfinished words, to detect disfluencies. However, those approaches only detect disfluency within the transcript while neglecting the location and duration of the speech disfluency, which plays an important role in the disfluency analysis, e.g. for the assessment of interlocutors' alignment in collaborative activities \cite{norman2022studying}. 

Recent work focuses on detecting speech disfluencies at the frame level to capture timing information. \cite{harvill22_interspeech, shonibare2022enhancing} investigate fine-tuning ASR models with disfluent dataset. \cite{kouzelis2023weaklysupervised} explores forced alignment that aligns the audio signal with its corresponding transcript by decoding with
Weighted Finite State Transducers (WFSTs) in alignment graph. \cite{lian2023unconstrained, lian2024hierarchical} hierarchically integrate transcription and detection modules. However, these works address predefined or restricted disfluency types and fail on the open-set disfluency detection for handling previously unseen types.

In this work, we propose a straightforward yet effective pipeline to augment ASR models by detecting open-set speech disfluency \footnote{https://github.com/Robin-Amann/bachelor-thesis}. As illustrated in Figure \ref{fig:Approach}, the pipeline consists of three hierarchical steps: transcription and feature extraction with the ASR model and a frame-wise feature extractor, transcript and speech alignment with a modified Connectionist Temporal Classification(CTC)-based approach from \cite{kurzinger2020ctc}, and alignment gaps classification for detecting the potential disfluencies. The contributions of this work are as follows:


\begin{itemize}
    \item We examine one state-of-the-art ASR model Whisper \cite{radford2023robust} on speech disfluency detection. The experimental results show the model achieves 22.54 Word Error Rate (WER) points on a conversational dataset, but only 56\% of speech disfluencies at the word level are correctly transcribed, and 73.77\% of untranscribed words are disfluencies. The results indicate that the ASR model performs poorly on disfluent speech, which is reasonable since ASR models are designed to produce fluent transcriptions for better readability. The finding highlights the importance of augmenting ASR models with disfluency detection capabilities.
    
    \item Aiming to detect the location and duration of speech disfluency, this work proposes a modified CTC-based forced alignment approach from \cite{kurzinger2020ctc} to effectively locate and capture speech disfluency. We compare the proposed approach with the popular CTC-based alignment \cite{kurzinger2020ctc} and Whisper's cross-attention alignment \cite{radford2023robust}, and show that the proposed algorithm clearly captures more untranscribed words than the others. 

    \item With the proposed forced alignment approach, we build an inference-only pipeline to augment ASR models with disfluency detection capability. The pipeline is flexible and can be adapted to any ASR model. In addition, the pipeline detects the alignment gaps containing disfluent speech, allowing the detection of the open-set disfluency beyond predefined types. With an alignment gap classification model, the pipeline achieves 81.62\% accuracy in identifying gaps containing speech, covering 74.13\% of all untranscribed words in the initial transcript.
    


\end{itemize}

%% file: sections/s2_disfluency_detection.tex
\section{Disfluency Detection}
\label{sec:forced_alignment}

\subsection{Augmentation pipeline}
The inference-only pipeline to augment ASR models with open-set disfluency detection consists of three hierarchical steps (\autoref{fig:Approach}). In the first step, the ASR system generates an estimated transcript, and a feature extractor model produces the frame-wise probability from speech. The ASR model could be the same as the frame-wise feature extractor models, such as Wav2Vec2 \cite{baevski2020wav2vec20frameworkselfsupervised}. But it can be any ASR model as the augmentation pipeline only needs its transcript. After that, the pipeline applies a modified (CTC)-based forced alignment algorithm, that is based on \cite{kurzinger2020ctc} with the above generations. The algorithm generates word-level timing information and the signal gaps between the word timesteps are recognized as potential instances of disfluent speech. In the end, a developed classification model is applied to identify alignment gaps containing disfluent speech or only silence. The classification results can be utilized for downstream tasks like second-step transcription and identifying disfluency types.

\subsection{Forced alignment}

As the key step to extract timing information, three forced alignment approaches are employed: the standard CTC forced alignment, a modified CTC forced alignment, and the cross-attention approach of Whisper, one of the SOTA ASR models, whose attention value exhibits a high degree of correlation with timestamps.

\subsubsection{Standard CTC Alignment}
\label{sec:standard_ctc}


CTC-based forced alignment is a popular approach to extracting timing information used in many speech recognition packages, such as ESPnet \cite{watanabe2018espnetendtoendspeechprocessing}, SpeechBrain \cite{ravanelli2021speechbraingeneralpurposespeechtoolkit} and Flashlight \cite{kahn2022flashlightenablinginnovationtools}. 
The alignment is calculated in three steps \footnote{https://pytorch.org/audio/stable/tutorials/forced\_alignment\_tutorial.html}:
\begin{enumerate}
    \item The audio is fed into a feature extraction model that is pre-trained with CTC to generate frame-wise label probability over the whole alphabet.
    \item From the probability, a trellis matrix is generated to represent the probability of labels occurring at each time frame.  The trellis at point $(t, j)$ (where $ 0 \leq t \leq T-1 \; and \; 0 \leq j \leq U-1 $) represents the maximal probability that the first $j-1$ labels of the transcripts are aligned to the first $t-1$ timeframes of the audio. The trellis is calculated in the log domain to avoid numerical instability. 
    
    The maximum probability that the first $j$ labels of the transcript are aligned at the timeframe $t$ is the maximum of 1) Stay on the label, which means the first $j$ labels of the transcript are already aligned at time $t-1$, and the alignment remains with the same label; 2) Switch to next label, which means that the first $j-1$ labels of the transcript are aligned at time $t-1$, and the alignment switches to label $j$ at time $t$. The calculation is as follows:
    \begin{equation}
    \begin{split}
     trellis[j, t] = max( trellis[j, t-1] \cdot prob[t, \epsilon], \\ trellis[j-1, t-1] \cdot prob[t, j])
     \end{split}
     \label{eq:trellis_ctc}
    \end{equation}

    where $trellis[j, t]$ and $prob[j,t]$ indicate the trellis value and the frame-wise probability value at time $t$ and label $j$, respectively.

    \item In the third step, the path with the highest probability in the trellis is traced back. This path begins at position $(0, 0) and ends at position (T, U)$ to encompass the entire audio and transcript.
    
\end{enumerate}




\subsubsection{Modified CTC Alignment}
\label{sec:modified_ctc}

\begin{figure}[h]
    \centering
    \includegraphics[height=5cm]{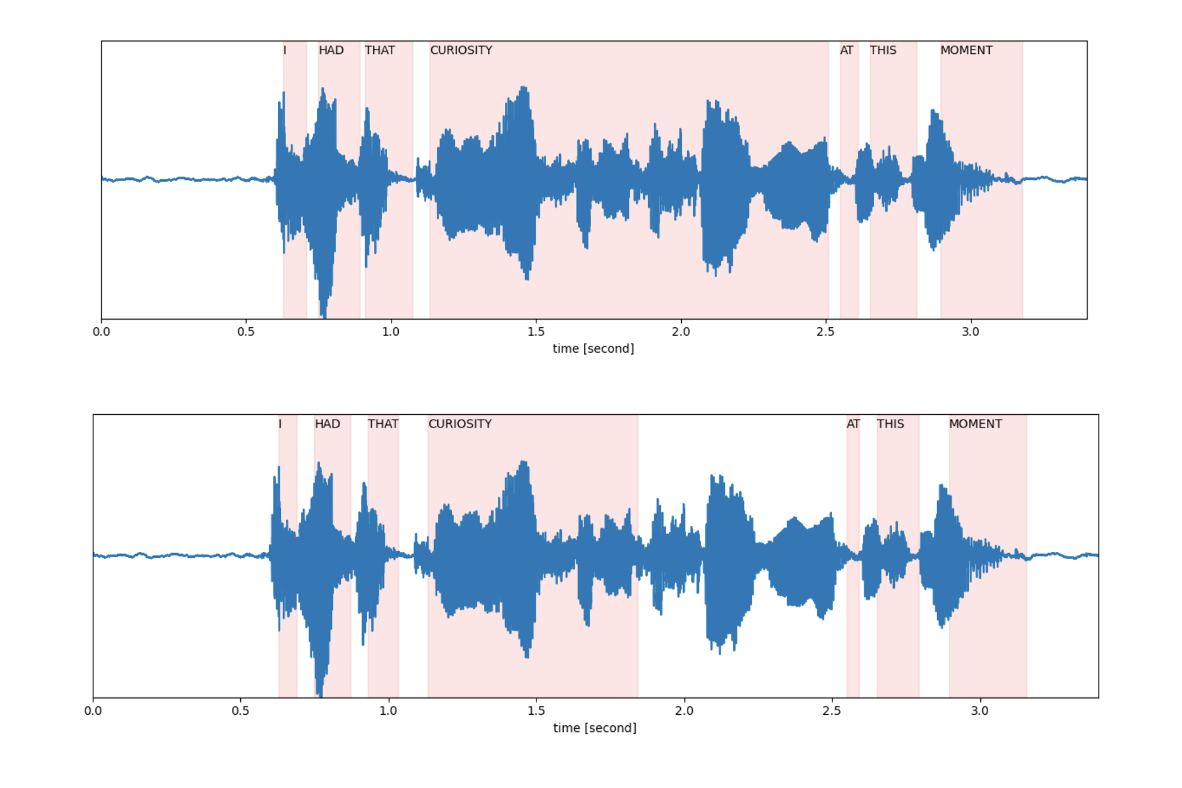}
    \caption{Alignments of generated transcription to speech signal with standard (upper) and modified (lower)  CTC forced alignments. The ASR prediction is: \textit{I had that curiosity at the moment}, while the manual transcript contains the disfluency: \textit{I had that curiosity \textbf{beside me} at the moment}.}
    \label{fig:ctc_alignments}
\end{figure}

Our preliminary experiments show that the standard CTC alignment struggles to generate correct information when the automated transcription does not include the disfluency. As \autoref{fig:ctc_alignments} shows, the manual transcript of this example contains the disfluency, which is removed by the ASR model for better readability. The standard CTC alignment extends the alignment around incomplete words, leading to inaccurate alignment and missing disfluency detection. This occurs because the standard CTC algorithm tends to align a word for a longer duration rather than to align silence where something is being said. In trellis generation (refer to \autoref{eq:trellis_ctc}), the emissions for a blank token in this part of the audio are very low, as something was spoken there.  As a consequence, gaps in the alignment of the transcript may not occur where they should, which is undesirable for this application.

To counteract this issue, we proposed the modified CTC alignment alignment to enable the model to detect the alignment gaps containing the speech of untranscribed disfluency. In trellis generation, the modified \autoref{eq:trellis_ctc_new} is applied \textbf{if the current label $j$ is a space token}. The space token infers a special label used to represent gaps or spaces between characters or tokens in the output space. With modification, the modified probability of staying on this label is the maximum of the probability acquired through the emissions and a predefined probability $c$. This modification incentivises the algorithm to extend silence between words to some extent.

 \begin{equation}
    \begin{split}
     trellis[j, t] = max(trellis[j, t-1] \cdot \\ \textbf{max}(prob[t, \epsilon], \textbf{c}), trellis[j-1, t-1] \cdot prob[t, j])
     \end{split}
     \label{eq:trellis_ctc_new}
    \end{equation}

As for the previous examples shown in \autoref{fig:ctc_alignments}, the modified CTC alignment correctly aligns the speech signals to the ASR prediction while capturing gaps that correspond to untranscribed disfluencies. However, it is also important to note that the modified alignment for, e.g., ``that" has also become shorter. This is because this modification encourages the algorithm to keep words short since an alignment containing much silence gives a better score. Setting the probability $c$ too high may result in words being too short overall. Therefore, we experiment with different predefined probabilities to choose the value carefully.

\subsubsection{Cross-attention Alignment}
\label{sec:cross_attention}

Whisper is trained in such a way that there exists a correlation between the cross-attention weights and the audio input. Consequently, the cross-attention weights and the output transcript allow for the calculation of alignment to the input audio through Dynamic Time Warping (DTW) \cite{JSSv031i07}. As implemented in the HuggingFace library \footnote{https://github.com/huggingface/transformers/blob/main/src/\newline transformers/models/whisper/generation\_whisper.py}, token-level timesteps are calculated using the encoder-decoder cross-attentions and DTW to map each output token to a position in the input audio.

\subsection{Alignment Comparison Metric}
\label{sec:metric}

Comparing the alignment approaches requires an automatic metric, which is not available. This work proposes a metric that evaluates the alignment between the manual and automatic transcripts by considering the position and length of aligned words. The correctly transcribed words are extracted using Levenshtein Alignment, and we denote $(s1, e1)$ and $(s2, e2)$ as the manual timing annotation and automatically aligned timing information.

After forced alignment,  timing information of the same word from manual annotation $A1$ and automated transcription $A2$. For each pair of words $(w1 , w2)$, the length, position and combined scores are calculated with Equations \ref{eq:Audio_Transcript_Alignmnet:Metric_position}, \ref{eq:Audio_Transcript_Alignmnet:Metric_length} and \ref{eq:Audio_Transcript_Alignmnet:Metric_combine}, respectively:

\begin{equation}
\label{eq:Audio_Transcript_Alignmnet:Metric_position}
m_p(w_1, w_2) =\frac{1}{ \mid \frac{p_1-p_2}{l_1} \mid +1 } 
\end{equation}
\begin{equation}
\label{eq:Audio_Transcript_Alignmnet:Metric_length}
m_l(w_1, w_2) =\frac{1}{ \mid \frac{l_1-l_2}{l_1} \mid +1 }    
\end{equation}

\begin{equation}
\label{eq:Audio_Transcript_Alignmnet:Metric_combine}
m(w_1, w_2) = \frac{1}{ \mid \frac{p_1-p_2}{l_1} \mid +1 } \cdot \frac{1}{ \mid \frac{l_1-l_2}{l_1} \mid +1 }    
\end{equation}

Where $w$ indicate a word with starting ($s$) and ending ($e$) time, and $p$ and $q$ indicate the position and length of the word calculated with $p=\frac{s+e}{2}$ and $l=\frac{e-s}{2}$. The final scores for the entire alignments are then computed by averaging the scores of individual words. The three scores range from 0 to 1, and a higher value indicates a better alignment.



\subsection{Gap classification}
The forced alignment detects the alignment gaps, while the gaps can contain speech or only silence. Accordingly, we propose a classification step to identify alignment gaps where disfluent speech may occur. 

Since the timing information of the disfluent speech is not available in the dataset, we define an alignment gap containing disfluent speech as one that covers at least one word. We define the coverage as the duration of the word has more than 50\% overlapping with the duration of the alignment gap. The 50\% overlap criterion strikes a balance: Considering only words completely within the gap would result in many gaps being deemed empty, despite there being speech intuitively present. Conversely, if a gap is considered non-empty as soon as a word is even partially within it, the transcription model may find it challenging to transcribe this word in the subsequent step.

Classification with all gaps is inefficient and might involve too small gaps due to alignment inaccuracy. Therefore, this work performs classification on gaps that exceed a minimum threshold length. The chosen minimum gap size should not be too small, as minimal inconsistencies in the alignment would become too noticeable. This would result in previously transcribed speech being transcribed again. On the other hand, the minimal gap should not be too large, as this could lead to overlooking too many gaps where untranscribed speech may be present. Based on preliminary experimental results, a gap size of 0.3 is selected as optimal.


%% file: sections/s3_dataset.tex
\section{Dataset}
\label{sec:dataset}
\subsection{Dataset Preparation}

As this work aims to detect the location and duration of speech disfluencies, the dataset must be composed of spontaneous speech with word-level timing annotation. Besides, the dataset must be large enough to act as training data for the classification model. Therefore, we use the Switchboard-1 Release 2 \footnote{https://catalog.ldc.upenn.edu/LDC97S62} and Treebank 3 \footnote{https://catalog.ldc.upenn.edu/LDC99T42} datasets. The Switchboard dataset consists of approximately 260 hours of telephone conversations with word-level timing information, and the Treebank 3 dataset adds the corresponding word-level disfluency annotation to the transcripts of the Switchboard dataset.

\subsection{Segmenting Audio}
\label{sec:Dataset_Preparation:Segmenting}

This dataset consists of recordings that are several minutes long. The long recordings hinder forced alignment performance, and the segmentation pre-processing is applied.

First, the audio file is segmented at all points where there is more than 5 seconds of silence. These points provide good places to divide the transcript without interrupting the speaker's flow of speech. However, there are still segments that remain several minutes long.

In the next step, as long as the segment is longer than 30 seconds, the largest gap between two words in the middle of the transcript is sought, ensuring that it is at least 10 seconds away from the beginning and end of the segment. This ensures that the resulting segments are between 10 and 30 seconds long. Of course, it is also possible that shorter segments are created when splitting by 5-second pauses.

%% file: sections/s4_exp_results.tex
\section{Experiments and Results}
\label{sec:results}

\subsection{Experimental setups}
\label{sec:results_exp_setups}

This work experiments with the SOTA ASR model Whisper \footnote{https://huggingface.co/openai/whisper-large-v3} to augment disfluency detection ability. As for frame-wise feature extractor, this work selects Wav2Vec2 \footnote{https://huggingface.co/facebook/wav2vec2-base-960h} which is fine-tuned on English ASR. The ASR model can be the same as the feature extractor model in pipeline design, but we select Whisper as the ASR model because this pipeline supports augmenting any ASR model, and Whispers is a stronger ASR model in terms of accuracy and robustness for this dataset. We believe our approach is effective for ASR models providing more accurate predictions than Whisper. Besides, we acknowledge that our approach may suffer performance degradation with ASR models yielding less accurate predictions.


\subsection{Are ASR models good at disfluency recognition?}
\label{sec:results_original_transcription}
For the initial transcription of the audio, we employ the Whisper model on the dataset and achieve 22.54 WER points. To assess the percentage of fluent and not fluent speech transcribed, the manual and the automatic transcript are aligned using the operations obtained during the calculation of the WER. \autoref{tab:Results_and_Analysis:Audio_Transcription:Transcribed_Words} illustrates the number of words from the manual transcript that were correctly transcribed, incorrectly transcribed and not transcribed, each annotated as fluent or not fluent speech. 

\begin{table}[ht]
    \centering
    \begin{tabular}{cccc}
        \hline
         & Correctly  & Incorrectly & Untranscribed\\
        & Transcribed & Transcribed & \\
        \hline
        Fluent & 895,474 & 45,823 & 31,931 \\ 
        Disfluent & 136,718 & 15,242 & 89,799 \\
        \hline
    \end{tabular}
    \caption{Fluent and disfluent transcribed words.}
    \label{tab:Results_and_Analysis:Audio_Transcription:Transcribed_Words}
\end{table}

It can be observed that approximately 74\% of all untranscribed words are labelled as speech disfluencies. Additionally, it is evident that 37\% of all speech disfluencies are not transcribed, 6\% are transcribed incorrectly and only 56\% are transcribed correctly, confirming the initial assumption that Whisper does not fully transcribe speech disfluencies.

\subsection{What parameter to use for modified CTC algorithm?}
\label{sec:results_probability_c}
This work proposes a modified CTC alignment algorithm to overcome the problems of the standard CTC alignment algorithm on disfluency recognition (Section \ref{sec:modified_ctc}). The modified algorithm involves a fixed probability \textit{c} as to incentive gap recognition in alignment.
The probability \textit{c} experimented with probability values -5, -4, -3, -2, -1, -0.5, -0.1, -0.01 on a par with the log probability scale in frame-wise acoustic probability. Setting as -0.01 is essentially 0 and the value above 0 makes no sense as for comparison with log probability. A higher value of probability $c$ indicates a higher chance to stay with the space token, corresponding to more and longer alignment gaps.



\begin{figure}[!h]
    \centering
    \includegraphics[width=9cm]{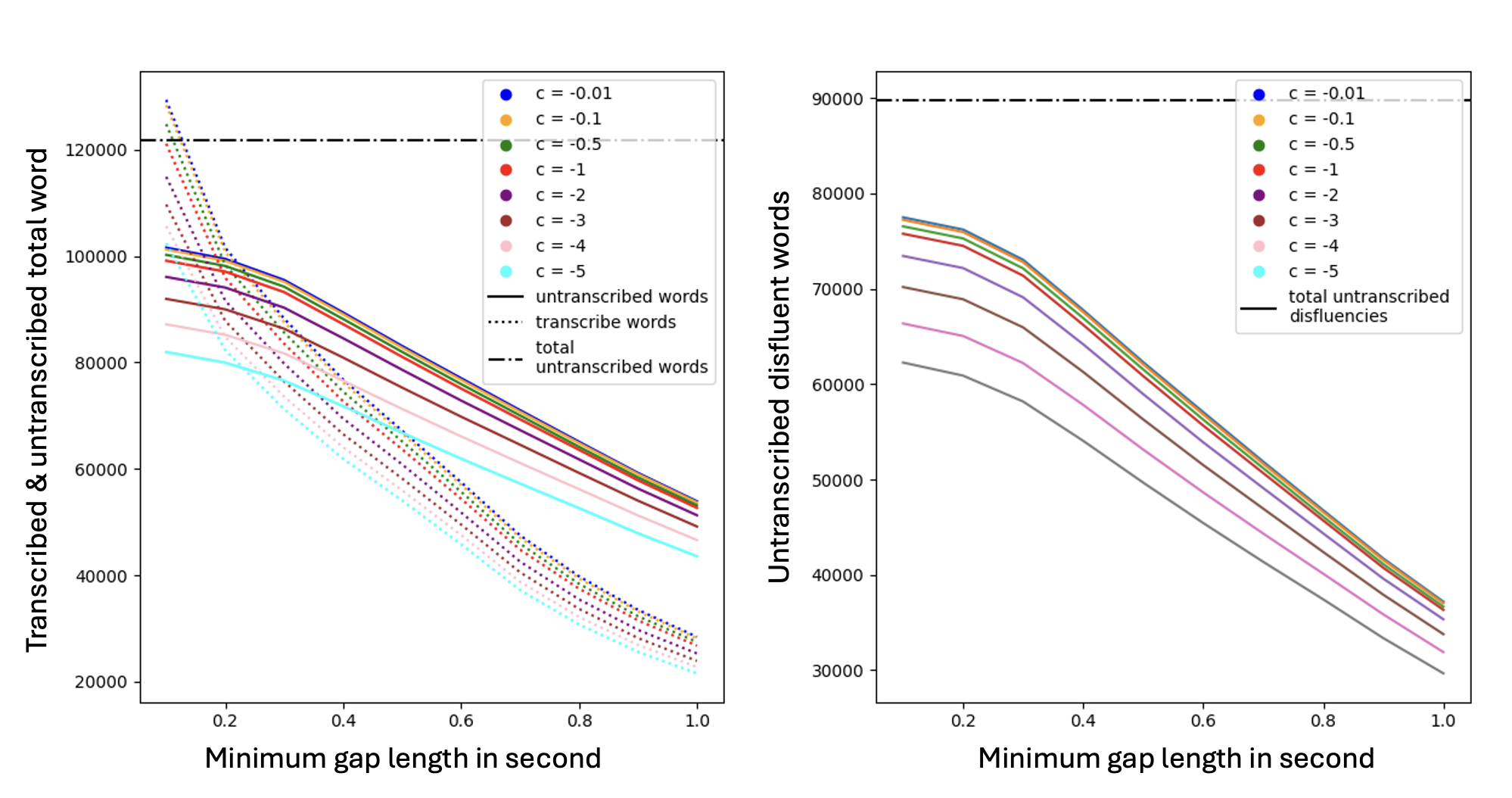}
    \caption{The number of words that are covered by modified alignments with different probability $c$ value. The left is for all words and the right is for the words next to the untranscribed words in the manual transcript.}
    \label{fig:c_value}
\end{figure}

This work counts the number of words that are covered by the alignment gaps to evaluate alignment performance. We evaluate the alignment approaches on all words in the manual transcription to show the general alignment performance, and we also evaluate them on only the words next to the untranscribed words to show the performance specifically on disfluent speech. Note that the untranscribed words are determined with Levenshtein Alignment, same as Section \ref{sec:metric}. As \autoref{fig:c_value} shows, significantly more untranscribed words are reachable with a default probability of -0.01 than -5.

Besides, we evaluate using the proposed alignment scores. As shown in \autoref{tab:Results_and_Analysis:Audio_Transcript_Alignment:Transcribed_Words}, the combined alignment scores for all words show no significant difference, but the score for words around the untranscribed words improves clearly with decreasing the probability value of $c$, which is consistent to \autoref{fig:c_value}. Therefore, the -0.001 is chosen for the modified CTC algorithm in later experiments.

\begin{table}[h]
    \centering
    \begin{tabular}{ccc}
        \hline
        Probability value & All words & \makecell{Words around \\ untranscribed words} \\
        \hline
        -0.01 & 0.5893 & 0.5628  \\
        -0.1  & 0.5896 & 0.5627\\
        -0.5  & 0.5906& 0.5627 \\
        -1  & 0.5917 & 0.5625\\
        -2 & 0.5932& 0.5600  \\
        -3 & 0.5937 & 0.5550 \\
        -4  & 0.5937 & 0.5484\\
        -5  & 0.5932 & 0.5407 \\
        \hline
    \end{tabular}
    \caption{Experimental results on modified CTC alignment with different predefined probability using the combined alignment score.}
    \label{tab:Results_and_Analysis:Audio_Transcript_Alignment:Transcribed_Words}
\end{table}

\subsection{Which forced alignment algorithm works better?}
\label{sec:results_alignment_comparsion}
The alignments calculated by the standard CTC alignment, the modified CTC algorithm and cross-attention are compared with the proposed evaluation metric.  Same as Section \ref{sec:results_probability_c}, we evaluate the performance for all words as well as only for the words around the untranscribed speech. But here we calculate the proposed alignment metrics of the position, length and the combined scores.

As \autoref{tab:methods_all_words}  shown, the modified CTC algorithm outperforms the others in all aspects, except the length for all words where the standard CTC is slightly better. For the alignment performance in the presence of untranscribed speech, the modified CTC algorithm shows a clear performance gain.

\begin{table}[h]
    \centering
    \begin{tabular}{cccc}
        \hline
        &  Cross Attention & CTC &  Modified CTC\\
        \hline
            \multicolumn{4}{l}{All words} \\ \hline
        Position&0.5941 &  0.7465         &  \textbf{0.7702}  \\
        Length&0.6855 & \textbf{0.7755} &           0.7612  \\
        Combined   &   0.4359 &          0.5880 &   \textbf{0.5893} \\
        \hline
         \multicolumn{4}{l}{Words around untranscribed words} \\ \hline
         Position                    &     0.5138   &   0.7177 &  \textbf{0.7619} \\
        Length                      &     0.6051   &   0.7376 &   \textbf{0.7555} \\
        Combined &     0.3508   &   0.5493 &   \textbf{0.5802} \\ \hline
        
    \end{tabular}
    \caption{Comparison of forced alignment algorithms with evaluation of alignment metric. \textit{Position} and \textit{Length} indicate the individual score, and \textit{Combined} indicates the score considering position and length (Section \ref{sec:metric}).}
    \label{tab:methods_all_words}
\end{table}

\begin{figure}[h!]
    \centering
    \includegraphics[width=6cm]{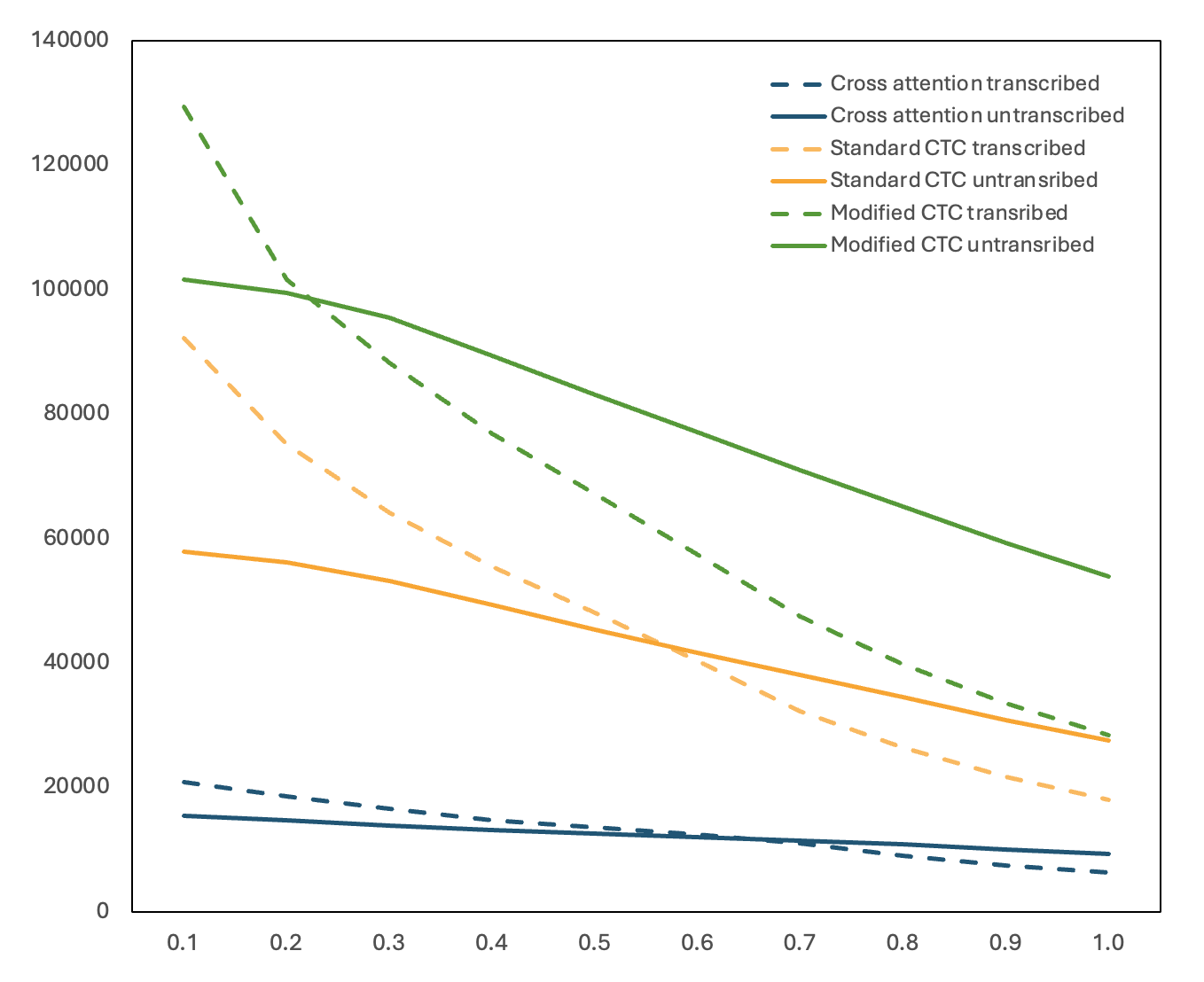}
    \caption{Untranscribed and transcribed words covered by three forced alignment approaches.}
    \label{fig:Methods_Comparison_Reachable}
\end{figure}

Besides evaluating the metric scores, this work also counts how many untranscribed words are covered.  \autoref{fig:Methods_Comparison_Reachable} shows for each presented alignment method the number of untranscribed words and already transcribed words within a gap for various minimum gap sizes. As can be seen, the modified CTC algorithm recognized many more untranscribed words than other algorithms. Specifically, with a total amount of 121,738,  modified CTC, standard CTC and cross-attention cover 81.69\%, 46.10\% and 12.02\% untranscribed words, respectively. Therefore, the modified CTC alignment is chosen for further analysis. 

\subsection{How to build disfluency classification model?}

The forced alignment brings gaps between the words of transcription, and the next question comes as to how to classify the gaps as containing speech or empty. This work proposes to build a classification model and train it with a dataset tailored for the pipeline with the gaps. 

With the combined dataset, we build the datasets consisting of extracted gaps. The gap is labelled as ``gap contains speech'' if at least one word from the manual transcript falls into this gap. Otherwise, they were labelled as ``gap is empty''. We use the modified CTC algorithm for alignment. After shuffling, we select 80\% of the gaps for training data, and the rest for test data (\autoref{tab:Untranscribed_Speech_Detection:Classification_Model:Dataset_Split}).

\begin{table}[h]
    \centering
    \begin{tabular}{cccc}
        \hline
          & Total & Containing Speech & Empty \\
        \hline
        Training & 220,344 & 101,207 & 119,137 \\
        Test & 40,980 & 19,651 & 21,329 \\
        \hline
    \end{tabular}
    \caption{Statistics of the gaps classification dataset}
    \label{tab:Untranscribed_Speech_Detection:Classification_Model:Dataset_Split}
\end{table}

With the above dataset, this work builds a classification model by fine-tuning a wav2wec2 model in conjunction with a classification head. The classification head consists of a linear layer projecting the output of Wav2Vec2 onto the predefined classes: an empty gap and a gap with speech. With the evaluation of the test split, the classification model achieved an accuracy of 81.62\%, a precision of 86.14\%, a recall of 74.80\%, and an F1-score of 80.07\%.




\subsection{How effective is the disfluency detection pipeline?}

Knowing the proportion of all gaps successfully classified does not provide information about the proportion of untranscribed words successfully classified. Therefore, we count the number of transcribed and untranscribed words that are classified and covered in the gaps, classified but not covered in gaps and not classified by the gap classification model.


As \autoref{tab:pipeline_performance} shows, 15,478 out of a total of 20,880 untranscribed words are covered by the detected alignment gaps, leading to a detection rate of 74.13\%. However, 13,202 out of 168,256 already transcribed words are also labelled as predicted in the gaps, counting to a false detection rate of 8.6\%. The false detection rate indicates the risk of double transcription if a follow-up re-transcription was carried on.

\begin{table}[h]
\centering
\begin{tabular}{ccc} \hline
                     & Transcribed & Untranscribed \\ \hline
Classified and covered     & 13,202      & 15,478        \\
Classified but uncovered           & 153,975     & 3,952       \\ 
Not classified & 1,079       & 1,450        \\ 

\hline
\end{tabular}
\caption{Pipeline performance evaluation on the type of words covered by the gaps.}
\label{tab:pipeline_performance}
\end{table}

%% file: sections/s5_conclusion.tex
\section{Conclusion}
\label{sec:conclusion}

In this work, we propose an inference-only pipeline to augment any ASR model with open-set disfluency detection. We reveal the current ASR models struggle to transcribe speech disfluency. To tackle this issue, we propose a modified CTC forced alignment algorithm to recognize the location and duration of speech disfluencies. We show the effectiveness of this approach by comparing it with popular forced alignment approaches in disfluency recognition. Additionally, we build a pipeline for disfluency detection and show that the approach captures 74.13\% of the words that are not transcribed by the initial transcription. 

However, the disfluency detection performance is dependent on the ASR model performance. This is because that transcribed disfluencies will not be identified as disfluencies, as they are already aligned with the transcribed words.

